%% file: arxiv.tex
\documentclass[sigconf,nonacm,pdfa]{acmart}
\makeatletter
\let\@ACM@checkaffil\@empty
\makeatother

\usepackage[ruled,vlined,linesnumbered]{algorithm2e}
\SetAlgoNoEnd
\usepackage{amsmath}
\usepackage[most]{tcolorbox}
\usepackage{xcolor}
\definecolor{planninggreen}{RGB}{0,170,0}
\definecolor{promptbackground}{RGB}{247,247,247}
\usepackage{bm}

\usepackage{multirow}

\newcounter{examplectr}
\renewcommand{\theexamplectr}{\arabic{examplectr}}
\newtcolorbox{examplebox}[2][]{%
  enhanced,
  breakable,
  colback=gray!5,
  colframe=gray!160,
  fonttitle=\bfseries\small,
  title={%
    Prompt~\refstepcounter{examplectr}%
    \label{#1}\theexamplectr: #2%
  },
  boxrule=1pt,
  arc=0mm,
  left=3pt,
  right=3pt,
  top=1pt,
  bottom=1pt,
  before upper={\scriptsize},
  pad at break*=2pt,
}

\AtBeginDocument{%
  }

\begin{document}

\title{EvoTS-Agent: A Self-Evolving LLM Agent for Financial Time Series Change Point Detection}

\author{Lei Jiang}
\authornote{These authors contributed equally to this work.}
\affiliation{%
  \institution{Alan Turing Institute}
}
\email{ljiang@turing.ac.uk}

\author{Ye Wei}
\authornotemark[1]
\affiliation{%
  \institution{University of Oxford}
}
\email{ye.wei@imm.ox.ac.uk}

\author{Xinyu Xi}
\authornotemark[1]
\affiliation{%
  \institution{National University of Singapore}
}
\email{xinyu\_xi@u.nus.edu}

\author{Jordan Langham-Lopez}
\affiliation{%
  \institution{Alan Turing Institute}
}
\email{jlanghamlopez@turing.ac.uk}

\author{Yifan Bao}
\affiliation{%
  \institution{National University of Singapore}
}
\email{yifan\_bao@comp.nus.edu.sg}

\author{Raad Khraishi}
\affiliation{%
  \institution{NatWest AI Research}
}
\email{Raad.Khraishi@natwest.com}

\author{Yihao Ang}
\authornote{These authors are corresponding authors.}
\affiliation{%
  \institution{National University of Singapore}
}
\email{yihao\_ang@comp.nus.edu.sg}

\author{Anthony K. H. Tung}
\authornotemark[2]
\affiliation{%
  \institution{National University of Singapore}
}
\email{atung@comp.nus.edu.sg}

\author{Lukasz Szpruch}
\authornotemark[2]
\affiliation{%
  \institution{University of Edinburgh}
}
\email{l.szpruch@ed.ac.uk}

\author{Hao Ni}
\authornotemark[2]
\affiliation{%
  \institution{University College London}
}
\email{h.ni@ucl.ac.uk}

\input{sec/0_abstract}


\ccsdesc[500]{Computing methodologies~Machine learning}
\ccsdesc[500]{Computing methodologies~Artificial intelligence}
\ccsdesc[300]{Mathematics of computing~Time series analysis}
\ccsdesc[300]{Information systems~Data mining}

\keywords{Financial Time Series; Change Point Detection; LLM; Agents}

\maketitle

\input{sec/1_intro}
\input{sec/2_relatedwork}
\input{sec/3_prelimnary}

\input{sec/4_method}

\input{sec/5_experiment}
\input{sec/6_conclusion}
\input{sec/7_acknowledge}

\bibliographystyle{ACM-Reference-Format}
\bibliography{sample-base}


\end{document}

%% file: sec/0_abstract.tex
\begin{abstract}

Financial time series exhibit non-stationary and heterogeneous statistical properties, making change-point detection challenging because no single unsupervised algorithm performs consistently across assets and market regimes. Conventional workflows consequently depend heavily on expert-driven model selection, feature design, and hyperparameter tuning, limiting their scalability and adaptability. We propose EvoTS-Agent, a validation-guided self-evolving LLM agent for autonomous financial time-series change-point detection. EvoTS-Agent first performs curated exploratory data analysis to characterize dataset properties and initialize candidate detection models. It then evolves executable experiment trajectories through three complementary operators: \textit{Revision} exploits the current best solution, \textit{Alternative Strategy} explores fundamentally different modeling directions when progress stagnates, and \textit{Recombination} synthesizes complementary evidence from high-performing trajectories. Validation feedback guides trajectory evolution throughout the search, enabling the agent to adapt its detection pipeline to the statistical characteristics of each dataset while preserving reliable optimization. Experiments across four benchmark datasets demonstrate that EvoTS-Agent consistently outperforms existing LLM-based agents while maintaining a 100\% execution success rate across all evaluated backbone LLMs. 
\end{abstract}

%% file: sec/1_intro.tex
\section{Introduction}

Financial markets are dynamic systems whose statistical properties evolve over time in response to changing macroeconomic conditions, investor behavior, liquidity, policy interventions, and unexpected external events. These changes can manifest as shifts in return distributions, volatility, correlation structures, trading activity, or other market characteristics \cite{ang2012regime}. Identifying such structural transitions is important for risk management, portfolio allocation, fraud and market-manipulation detection, and quantitative decision-making. Change-point detection provides a principled approach for locating moments at which the data-generating process changes, enabling analysts to segment financial time series into statistically distinct regimes and identify potentially significant market events \cite{aminikhanghahi2017survey}.

Despite substantial progress in change-point detection \cite{kanrar2025model,wu2024unsupervised,londschien2023changeforest,el2024variational}, its application to financial data remains challenging.
Financial time series are typically noisy, non-stationary, heavy-tailed, and characterized by time-varying volatility and dependence structures. The nature and magnitude of structural changes might also vary considerably across assets, sampling frequencies, and market regimes. 
Consequently, no single change-point detection algorithm performs consistently well under all conditions \cite{truong2020selective}. A method effective for abrupt mean shifts in a univariate series may be unsuitable for gradual covariance changes in multivariate data, while an algorithm that performs well on synthetic benchmarks may degrade substantially when applied to real-world financial observations.

Selecting an appropriate detection pipeline therefore requires decisions at several levels, including the choice of algorithm, signal representation, preprocessing procedure, hyperparameters, detection thresholds, and post-processing rules. These decisions are often interdependent. For example, the usefulness of a kernel-based detector may depend on the scaling and dimensionality of the input, while the performance of a probabilistic method may be sensitive to assumptions about noise distributions or regime duration. In conventional workflows, these choices are made manually through domain expertise, repeated experimentation, and extensive parameter tuning, with model performance also being sensitive to the quality of the underlying data~\cite{maurino2025data}. Such workflows are difficult to scale across large collections of assets and datasets, and they may fail to adapt efficiently when the statistical characteristics of the observed market change.

\begin{figure}[t]
    \centering
    \includegraphics[width=\linewidth]{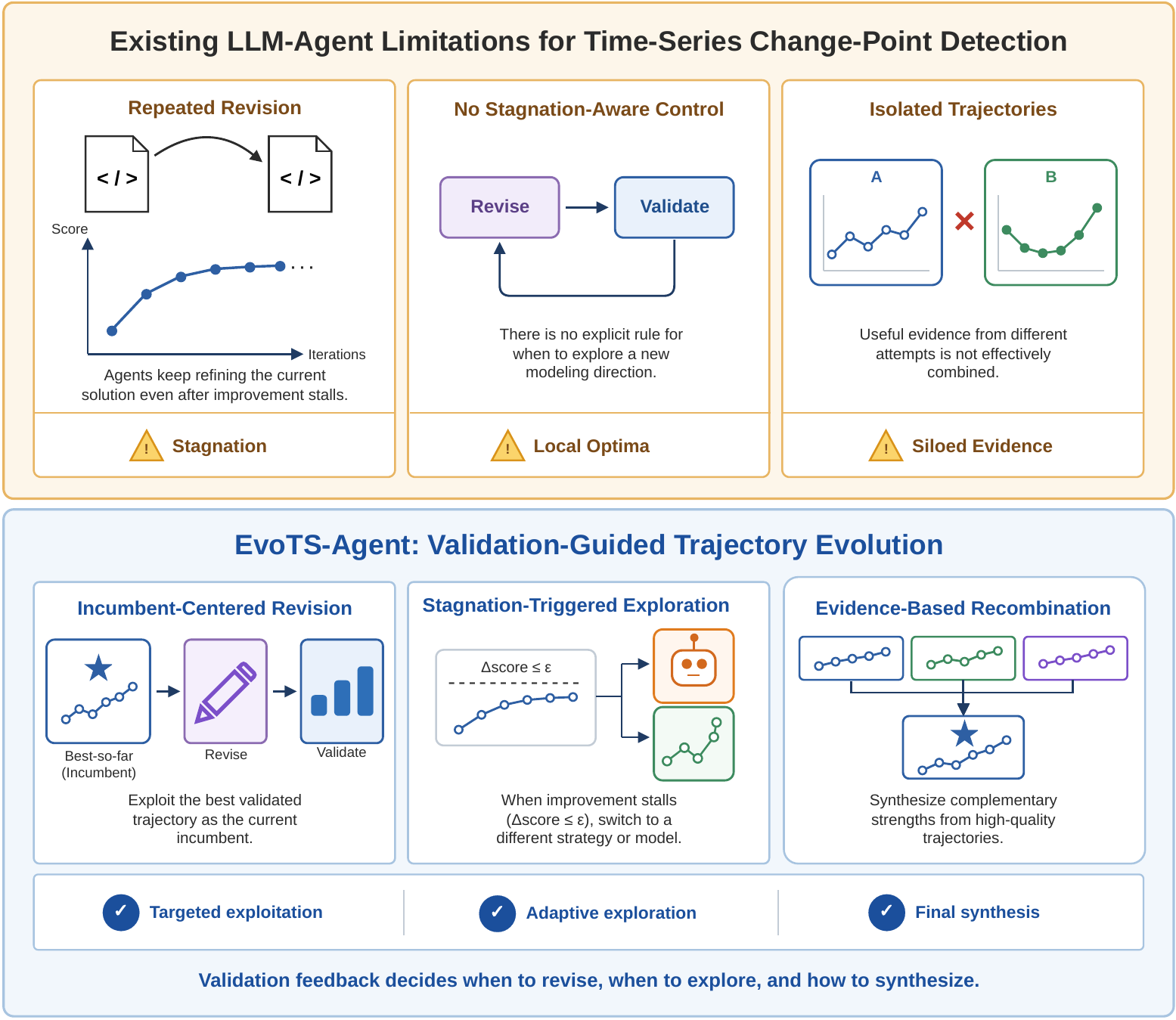}
    \caption{Motivation of EvoTS-Agent.}
    \label{fig:motivation}
\end{figure}

Large language models (LLMs) have recently shown promise as reasoning and coding components within autonomous agents \cite{tsgassist,guo2024ds,bao2026mosaic,zhang2026openfingym}. By combining natural-language planning, tool use, code generation, and environmental feedback, LLM agents can perform multi-step tasks that extend beyond a single model invocation. 
However, as shown in Figure~\ref{fig:motivation}, existing LLM-agent paradigms remain limited when applied to autonomous change-point detection. A predefined workflow may restrict the agent to a fixed sequence of operations even when validation evidence suggests that a different search strategy would be more appropriate. 
Retrieval-based systems \cite{guo2024ds,tsagent,tsautobox} can reuse previously successful solutions, but retrieved experiences may transfer poorly to datasets with different statistical properties. 
Moreover, simple iterative refinement \cite{guo2024ds,baek2024researchagent} typically improves a single solution through successive revisions, but lacks explicit optimization mechanisms for deciding when to exploit the current approach and when to explore fundamentally different alternatives. It also provides limited support for synthesizing complementary evidence from multiple successful experiments. Consequently, once local refinement ceases to produce meaningful improvement, the search may remain confined to a suboptimal modeling direction.

To address these challenges, we propose \textbf{EvoTS-Agent}, a self-evolving LLM agent for autonomous financial time-series change-point detection. EvoTS-Agent combines curated exploratory data analysis, model selection, executable experimentation, and validation-guided trajectory evolution within a closed-loop framework. The EDA stage characterizes dataset-specific properties and uses them to initialize a diverse set of candidate change-point detectors. EvoTS-Agent then improves executable experiment trajectories through three complementary operators. \textit{Revision} exploits the current incumbent by refining the best validated pipeline. When revision stagnates, \textit{Alternative Strategy} explores a fundamentally different modeling direction or an untried EDA-recommended detector. In the final stage, \textit{Recombination} synthesizes complementary evidence from high-performing trajectories. An incumbent-preserving selection rule ensures that unsuccessful experiments do not degrade the best validated solution.

We use the term \textit{self-evolution} to describe this inference-time transformation and selection of experiment trajectories rather than any update to the underlying language model. Each trajectory records the experimental plan, executable implementation, validation outcome, and evolutionary lineage, allowing subsequent decisions to be guided by accumulated empirical evidence. 

The principal contributions of this work are as follows:
\begin{itemize}
    \item To the best of our knowledge, EvoTS-Agent is the first self-evolving agent to automate financial time series change point detection.
    \item We build a comprehensive model bank, containing a diverse collection of change-point detection methods with fundamentally different assumptions, enabling the agent to handle a broad spectrum of change-point scenarios.
    \item We design a curated EDA process for time-series change-point detection that characterizes dataset properties and leverages them to guide the LLM in selecting candidate detection models.
    \item We propose an adaptive evolutionary policy that performs incumbent-based revision by default, activates alternative strategies when revision stagnates, and reserves recombination for final evidence-based synthesis.

\end{itemize}

Through these contributions, EvoTS-Agent reframes autonomous change-point detection as an empirical search over executable scientific experiments. Rather than relying on a fixed algorithm or a static agent workflow, the framework adapts its detection pipeline to the characteristics of each dataset and to the evidence accumulated during execution. This provides a foundation for scalable, transparent, and validation-driven financial change point detection.

%% file: sec/2_relatedwork.tex
\section{Related Work}
\subsection{Change Point Detection}
Change-point detection methods differ fundamentally in what they assume to be stable within a regime. The choice of assumption largely determines which types of structural changes can be detected effectively and, consequently, which application scenarios a method is best suited for \cite{truong2020selective}. Classical statistical methods assume that observations within each segment share constant distributional parameters, such as the mean, variance, or likelihood, and identify change points through optimal segmentation \cite{killick2012pelt,scott1974cluster,keogh2001segmentation} or Bayesian inference \cite{fearnhead2006bayesian}. More general nonparametric approaches instead assume that the entire data distribution remains unchanged within a regime, detecting changes using measures such as maximum mean discrepancy \cite{gretton2012kernel} or ensemble-based statistics, enabling greater flexibility for multivariate data \cite{ londschien2023changeforest}. For high-dimensional and complex time series, recent deep learning methods assume that regime changes are more distinguishable in a learned latent representation than in the original observation space, with approaches such as KL-CPD leveraging representation learning to capture subtle structural changes \cite{chang2019klcpd}. Other methods characterize regimes through spectral properties, detecting changes in frequency-domain behavior that may not be evident in the time domain \cite{preuss2015spectral}.

\subsection{LLM-based Agents}

LLM-based agents increasingly automate data-science workflows by combining planning, code generation, tool use, and execution feedback \cite{yang2026time,aggarwal2025information}. ReAct~\cite{yao2022react} establishes a general agentic paradigm by interleaving reasoning and acting, enabling LLMs to iteratively interact with external environments. Building on this paradigm, DS-Agent~\cite{guo2024ds} uses case-based reasoning to retrieve and adapt prior solutions, while ResearchAgent~\cite{baek2024researchagent} iteratively develops research ideas and experimental plans using literature retrieval and reviewer feedback. TS-Agent~\cite{tsagent} structures financial time-series modelling into model selection, code refinement, and fine-tuning, whereas MOSAIC~\cite{bao2026mosaic} grounds workflow construction in retrieved cases and reusable modelling modules through an intermediate blueprint representation. These methods improve automation through retrieval, modular orchestration, or repeated refinement, but generally follow a predefined workflow or continue revising the current solution without explicitly adapting the search policy.

SE-Agent~\cite{guo2026se} is most closely related to our work, as it evolves agent trajectories through revision, recombination, and refinement. However, SE-Agent primarily optimizes reasoning trajectories for software-engineering problem solving. EvoTS-Agent instead maintains validated executable experiment trajectories that jointly record the experimental plan, model and transformation choices, implementation changes, execution feedback, validation performance, and lineage. More importantly, EvoTS-Agent uses validation outcomes to control the evolutionary process: it revises the incumbent by default, activates an alternative strategy when revision stagnates, and performs evidence-based recombination for final synthesis. Thus, empirical feedback determines not only which solution is retained, but also how the subsequent search is conducted.

%% file: sec/3_prelimnary.tex
\section{Preliminaries and Problem Setup}
\label{sec:preliminaries}

\paragraph{Financial Time-Series Change-Point Detection Environment.}
We formulate autonomous financial change-point detection as an empirical optimization problem, in which an LLM agent iteratively proposes, executes, and validates candidate detection pipelines to maximize validation performance. Given a financial time series
\begin{equation}
    \bm{X}
    =
    (\bm{x}_1,\bm{x}_2,\ldots,\bm{x}_N),
    \qquad
    \bm{x}_i \in \mathbb{R}^{d},
\end{equation}
the objective is to identify a set of structural-change locations
\begin{equation}
    \widehat{\mathcal{B}}
    =
    \{\hat{b}_1,\hat{b}_2,\ldots,\hat{b}_m\},
    \qquad
    \hat{b}_j \in \{1,\ldots,N\},
\end{equation}
that approximates the reference boundary set $\mathcal{B}$. Here, $N$ is the sequence length and $d$ is the number of observed variables.

Although the underlying change-point detectors operate in an unsupervised manner, EvoTS-Agent performs model selection and trajectory evolution using validation feedback. Consequently, reference boundaries are available only on the validation split for evaluating experimental configurations, while test-set annotations and metrics remain hidden throughout the optimization process.

To formalize the optimization process performed by EvoTS-Agent, we model the interaction between the agent and the experimentation environment as

\begin{equation}
    \mathcal{E}
    =
    (\mathcal{U},\mathcal{X},\mathcal{A},F,Q),
\end{equation}
where $\mathcal{U}$ is the space of financial change-point detection tasks, $\mathcal{X}$ is the experimental state space, and $\mathcal{A}$ is the set of actions available to the agent. An experimental state $x \in \mathcal{X}$ contains the dataset profile, available and previously attempted models, executable scripts, validation observations, trajectory pool, and current incumbent.

The action space includes exploratory data analysis, candidate-model selection, experiment planning, script modification, execution, validation, trajectory revision, alternative-strategy generation, and trajectory recombination. The transition function
\begin{equation}
    F:
    \mathcal{X}
    \times
    \mathcal{A}
    \rightarrow
    \mathcal{X}
\end{equation}
maps the current experimental state and an agent action to a new state. The evaluation function
\begin{equation}
    Q:
    \mathcal{X}
    \times
    \mathcal{U}
    \rightarrow
    \mathbb{R}
\end{equation}
assigns a validation score to an executed experiment. For change-point detection, the primary score is boundary-aware validation F1:
\begin{equation}
    q_k
    =
    Q(\tau_k,u)
    =
    \operatorname{F1}_{\mathrm{val}}(\tau_k,u).
\end{equation}
Additional measures, such as Hausdorff distance, are retained as diagnostic evidence. In the current implementation, however, incumbent selection is determined by the primary scalar score $q_k$.

\paragraph{Experiment Trajectories.}
The central object in EvoTS-Agent is an executable experiment trajectory. Rather than storing only an LLM reasoning trace, each trajectory records the experimental evidence generated during one executable experiment, including its implementation, validation outcome, and evolutionary context.

The trajectory generated at optimization step $k$ is
\begin{equation}
    \tau_k
    =
    \left(
        P,\,
        op,\,
        \pi_k,\,
        h_k,\,
        M_k,\,
        S_k,\,
        q_k,\,
        L_k,\,
        D_k,\,
        a_k,\,
        z_k
    \right),
\end{equation}
where $P$ is the parent-trajectory set; $op$ is the selected self-evolution operation; $\pi_k$ is the evolved single-trial experiment plan; $h_k$ is an LLM-generated summary of the implemented experiment; $M_k$ is the executed model; $S_k$ is the resulting executable script; $q_k$ is the validation score; $L_k$ is the sanitized execution log; $D_k$ is the implemented code difference; $a_k \in \{\textsc{Accept},\textsc{Reject}\}$ is the incumbent-selection decision; and $z_k \in \{\textsc{True},\textsc{False}\}$ indicates whether a revision has stagnated.

All recorded trajectories are maintained in the trajectory pool
\begin{equation}
    \mathcal{T}
    =
    \{\tau_1,\tau_2,\ldots,\tau_k\}.
\end{equation}
The parent set $P$ gives each trajectory an explicit lineage. Consequently, $\mathcal{T}$ represents a directed experimental search graph rather than an unstructured conversational history.

%% file: sec/4_method.tex
\begin{figure*}[t]
    \centering
    \includegraphics[width=\linewidth]{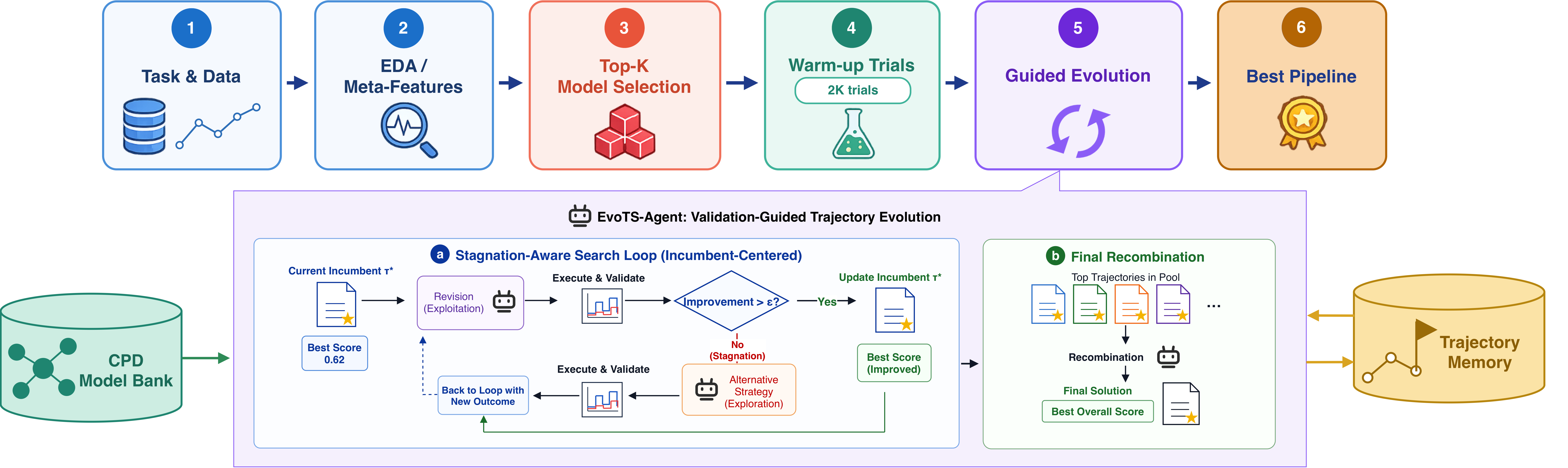}
    \caption{Orchestration of EvoTS-Agent.}
    \label{fig:workflow}
\end{figure*}

\section{EvoTS-Agent}

Figure~\ref{fig:workflow} illustrates the overall orchestration of EvoTS-Agent. Given a financial change-point detection task, the agent first performs a curated exploratory data analysis (EDA) to characterize the structural properties of the time series and extract dataset meta-features. Together with the raw time-series visualization, these meta-features are provided to the LLM, which selects the top-$K$ candidate change-point detection models from the model bank that are most suitable for the current task. During the warm-up stage, each selected model undergoes an initial implementation followed by a single revision, producing two executable experiment trajectories per model. Consequently, a total of $2\times K$ validated trajectories are generated and stored in the experiment trajectory memory. The trajectory achieving the highest validation score becomes the incumbent and serves as the starting point for subsequent optimization.

During the optimization stage, EvoTS-Agent iteratively improves the incumbent trajectory through three trajectory-level evolution operators. Under normal circumstances, the agent performs Revision, which refines the incumbent based on recent successful trajectories stored in memory. If the revision fails to produce meaningful validation improvement, the agent switches to Alternative Strategy, encouraging exploration of a substantially different modeling direction or an untried EDA-recommended model while retaining knowledge from previous attempts. At the final optimization step, the agent performs Recombination, synthesizing complementary strengths from multiple high-performing trajectories to generate the final experiment. Every executed experiment, including its implementation, validation result, code modifications, and summarized rationale, is recorded in the trajectory memory, allowing future decisions to leverage accumulated experimental evidence while preserving the best validated solution throughout the search. 


\subsection{EDA-based Model Selection}
Before optimization, the agent performs lightweight EDA to guide baseline model selection. A reproducibly sampled time series is summarized using temporal, spectral, and change-sensitive features, including lag-1 autocorrelation, trend strength, nonstationarity, spectral concentration, periodicity, and local mean and variance discrepancies. Local mean and variance discrepancies are designed to characterize the type and magnitude of possible structural changes by scanning adjacent windows around candidate boundaries. Specifically, for a window width $w$, the local mean- and variance-discrepancy statistics at position $t$ are defined as:
\begin{equation}
\Delta_{\mu}(t)
=
\frac{
\left|
\mu\!\left(X_{t:t+w}\right)
-
\mu\!\left(X_{t-w:t}\right)
\right|
}{
\sigma_x+\epsilon
},
\end{equation}
and
\begin{equation}
\Delta_{\sigma}(t)
=
\frac{
\left|
\sigma\!\left(X_{t:t+w}\right)
-
\sigma\!\left(X_{t-w:t}\right)
\right|
}{
\sigma_x+\epsilon
}.
\end{equation}

The maximum values over all valid positions summarize the strengths of
abrupt mean and variance changes:
\begin{equation}
S_{\mu}=\max_t \Delta_{\mu}(t),
\qquad
S_{\sigma}=\max_t \Delta_{\sigma}(t).
\end{equation}

Dataset properties such as sequence length, dimensionality, and missing-value ratio are also recorded. Ground-truth change points and test-set metrics are excluded to prevent information leakage.

The resulting meta features, an unlabeled time-series visualization, task metadata, and candidate model descriptions are provided to an LLM-based selector. The selector chooses \(K\) primary models for warm-up experiments and up to two alternatives. Primary models initialize independent optimization trajectories, while alternatives may be considered later if validation performance stagnates. Thus, EDA efficiently narrows the model search space while leaving final model selection to validation-based evaluation.

\subsection{Trajectory-guided Evolution}

\begin{algorithm}[h]
\caption{Trajectory-Guided Self-Evolution}
\label{alg:cpd-se-self-evolution}

\KwIn{Incumbent $(M^*,S^*,q^*)$, trajectory pool $\mathcal{T}$,
budget $\lambda$, threshold $\epsilon$}
\KwOut{Best validated model and script $(M^*,S^*)$}

\For{$k \gets 1$ \KwTo $\lambda$}{
    $\tau^* \gets \textsc{Incumbent}(\mathcal{T})$;
    $q_{\mathrm{prev}} \gets q^*$\;
    $C \gets \textsc{CurrentContext}
    \bigl(S^*,\textsc{RecentRelevant}(\mathcal{T},M^*,4)\bigr)$\;

    \uIf{$k=\lambda$}{
        $op \gets \textsc{Recombine}$\;
        $P \gets
        \textsc{SelectStrongTrajectories}(\mathcal{T},M^*)
        \cup\{\tau^*\}$\;
        $E \gets C\cup P$\;
    }
    \uElseIf{$\exists$ unused stagnated revision $\tau_s$ for $M^*$}{
        $op \gets \textsc{AlternativeStrategy}$\;
        $P \gets \{\tau^*,\tau_s\}$;
        $E \gets C\cup\{\tau_s\}$\;
    }
    \Else{
        $op \gets \textsc{Revision}$\;
        $P \gets \{\tau^*\}$;
        $E \gets C\cup P$\;
    }

    $\pi_k \gets \textsc{EvolvePlan}(op,E,M^*,S^*)$\;
    $(M_k,S_k,q_k,L_k,D_k)
    \gets \textsc{ExecuteAndValidate}(\pi_k,S^*)$\;

    \uIf{$q_k$ is valid and $q_k>q_{\mathrm{prev}}$}{
        $(M^*,S^*,q^*) \gets (M_k,S_k,q_k)$;
        $a_k \gets \textsc{Accept}$\;
    }
    \Else{
        $a_k \gets \textsc{Reject}$\;
    }

    $z_k \gets
    (op=\textsc{Revision})\land
    \bigl(q_k\text{ is invalid}\lor
    q_k-q_{\mathrm{prev}}\leq\epsilon\bigr)$\;

    $h_k \gets \textsc{SummarizeExperiment}(\pi_k,D_k)$\;

    $\tau_k \gets
    \textsc{RecordTrajectory}
    (P,op,\pi_k,h_k,M_k,S_k,q_k,L_k,D_k,a_k,z_k)$\;

    $\mathcal{T}\gets\mathcal{T}\cup\{\tau_k\}$\;
}

\Return{$(M^*,S^*)$}\;
\end{algorithm}


At the beginning of each optimization step, the agent retrieves the incumbent trajectory together with several recent relevant trajectories. These trajectories and the incumbent executable script form the planning context. Depending on the optimization state, the agent selects one of three trajectory-level operators—Revision, Alternative Strategy, or Recombination—and generates the next executable experiment plan using the current context and evolutionary evidence. The detailed control flow is given in Algorithm~\ref{alg:cpd-se-self-evolution}.


\textbf{Revision.} During ordinary optimization, the incumbent trajectory serves as the parent of the next experiment. The agent reflects on the incumbent script together with recent successful trajectories to produce exactly one executable modification. Revisions may alter the time-series representation, detector configuration, hyperparameters, or post-processing procedure, while preserving the overall experimental objective.

\textbf{Alternative Strategy.}
After each revision, the agent compares the resulting validation score with that of the incumbent. When the improvement is smaller than a predefined threshold (or the experiment fails to produce a valid score), the revision is marked as stagnant. At the next non-final iteration, the agent performs Alternative Strategy instead of another Revision. The stagnant trajectory is treated as negative evidence that discourages repeating the same search direction, while the incumbent script remains as the starting point for implementation. This encourages exploration of orthogonal modeling choices or previously untried EDA-recommended models.


\textbf{Recombination.} During the final optimization iteration, Recombination replaces both Revision and Alternative Strategy. Rather than extending a single trajectory, the agent synthesizes complementary components from multiple high-performing trajectories while retaining the incumbent script as the executable starting point. This allows the final experiment to integrate successful ideas discovered throughout the search.

\textbf{Incumbent Selection.} Every generated experiment is executed and evaluated on the validation set. The incumbent is updated only when the new experiment produces a valid validation score that is strictly better than the current incumbent. Otherwise, the incumbent is preserved. Acceptance and stagnation are intentionally independent: a small positive improvement is accepted because it improves the incumbent, yet it is still marked as stagnant if the improvement does not exceed the meaningful-improvement threshold. Consequently, the improved experiment becomes the new incumbent while simultaneously triggering Alternative Strategy in the following non-final iteration.

%% file: sec/5_experiment.tex
\section{Experiment}

\input{tables/agent_comparison}

\begin{figure}[t]
  \centering
  \includegraphics[width=0.5\textwidth]{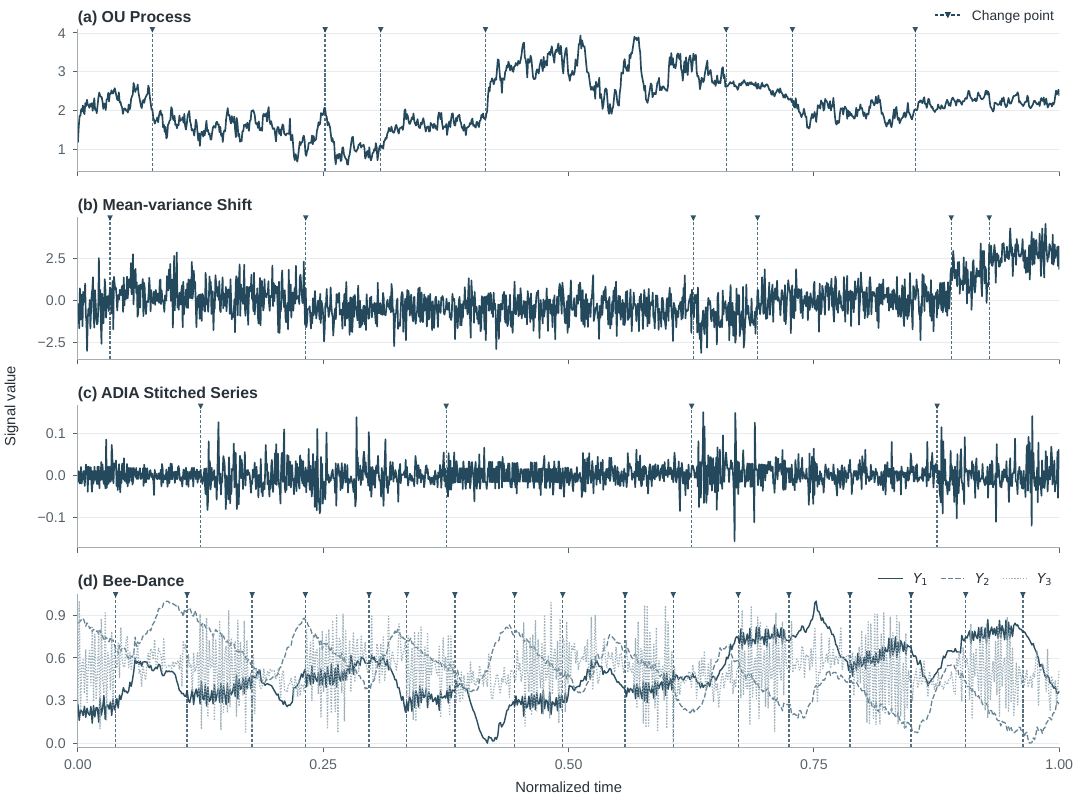}
  \caption{Visualization of raw input data. One representative sample from each dataset: (a) OU process, (b) Mean-variance Shift, (c) ADIA, and (d) Bee-Dance.}
  \label{fig:data}
\end{figure}

\textbf{Datasets.} We evaluate EvoTS-Agent on four complementary benchmarks designed to assess both controlled and real-world change-point detection capabilities. Figure~\ref{fig:data} shows a representative sample from each dataset. The first is a synthetic Piecewise \textbf{Ornstein Uhlenbeck (OU)} dataset that models regime shifts in mean-reverting financial processes through changes in the underlying OU parameters, providing a realistic approximation of assets such as credit spreads. The second is a synthetic \textbf{Mean+Variance Shift} dataset, where each regime is generated from a Gaussian distribution with segment-specific mean and variance, enabling evaluation under simultaneous changes in level and volatility that resemble transitions between calm and stressed market conditions. The third dataset is derived from the \textbf{ADIA} Lab Structural Break Challenge~\cite{adialab_structural_break_challenge_2025}, which consists of univariate time series annotated with a single structural break separating pre- and post-break regimes. Since each ADIA sequence contains only a single structural break, we construct longer evaluation streams by extracting fixed-width windows surrounding each annotated break and grouping windows with similar mean-shift magnitudes and volatility transitions. Windows within each group are then concatenated to produce multi-break sequences that emulate a continuous stream of market regime changes, in which periods of low and high volatility alternate across different assets or historical episodes. Although the stitched sequences are not chronological histories of individual financial instruments, they preserve realistic local dynamics around structural breaks and provide a challenging benchmark for autonomous change-point detection under successive volatility regime transitions. Finally, we include the \textbf{Bee-Dance} \cite{oh2008learning} dataset, a widely used real-world benchmark in change-point detection, in which the objective is to identify transitions between behavioral stages of a honey bee's waggle dance from motion trajectories. Together, these datasets cover controlled parameterized regime changes, compound distributional shifts, and noisy real-world signals, providing a comprehensive evaluation of the robustness and generalization ability of our agent. Following \cite{lai2018modeling,chang2019klcpd}, each dataset is split chronologically into training (60\%), validation (20\%), and test (20\%) sets.

\textbf{Model Bank.} Our model bank integrates eight complementary change-point detection methods. PELT, Bottom-up, and Window are implemented using the corresponding estimators provided by the ruptures library~\cite{truong2018ruptures}\footnote{\url{https://github.com/deepcharles/ruptures}}: PELT performs exact penalized segmentation with pruning~\cite{killick2012pelt}, Bottom-up greedily merges neighboring segments~\cite{keogh2001segmentation}, and Window detects changes through local sliding-window discrepancies. The remaining methods include ChangeForest-RF and ChangeForest-KNN~\cite{londschien2023changeforest}, Bayesian offline change-point detection~\cite{fearnhead2006bayesian}, KL-CPD~\cite{chang2019klcpd}, and our windowed spectral-discrepancy baseline motivated by frequency-domain structural break detection~\cite{preuss2015spectral}.

\textbf{Baselines.} We compare our agents with TS-Agent~\cite{tsagent}, DS-Agent \cite{guo2024ds} and ResearchAgent~\cite{baek2024researchagent}. We use GPT-4o~\cite{openai2023gpt4}, GPT-5.4~\cite{openai2025gpt5}, Claude Sonnet-4.6~\cite{anthropic2025claude} and Sonnet-5~\cite{anthropic2026sonnet5} for comprehensive comparison. Note that we disable the thinking mode of Sonnet-5 to ensure a fair comparison with models that do not support explicit reasoning modes. 

\textbf{Evaluation Metrics.} We evaluate detection quality using four complementary metrics. \textbf{F1} is the primary metric and is computed using boundary-aware matching between predicted and ground-truth change points within a predefined $\pm10$-sample tolerance window, balancing precision and recall. \textbf{Precision} measures the proportion of detected change points that correspond to true change points, while \textbf{Recall} measures the proportion of ground-truth change points that are successfully detected. To evaluate localization accuracy, we additionally report the \textbf{Hausdorff Distance}, which measures the maximum temporal deviation between the sets of detected and ground-truth change points; lower values indicate more accurate boundary localization. Finally, \textbf{Success Rate} reports the percentage of runs that complete successfully without execution or runtime failures, reflecting the reliability of each autonomous agent in producing executable solutions.

\subsection{Change Point Detection}

\begin{figure}[htbp]
  \centering
\includegraphics[width=0.3\textwidth]{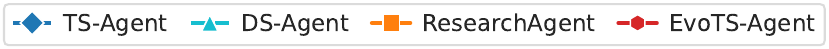}
\includegraphics[width=0.3\textwidth]{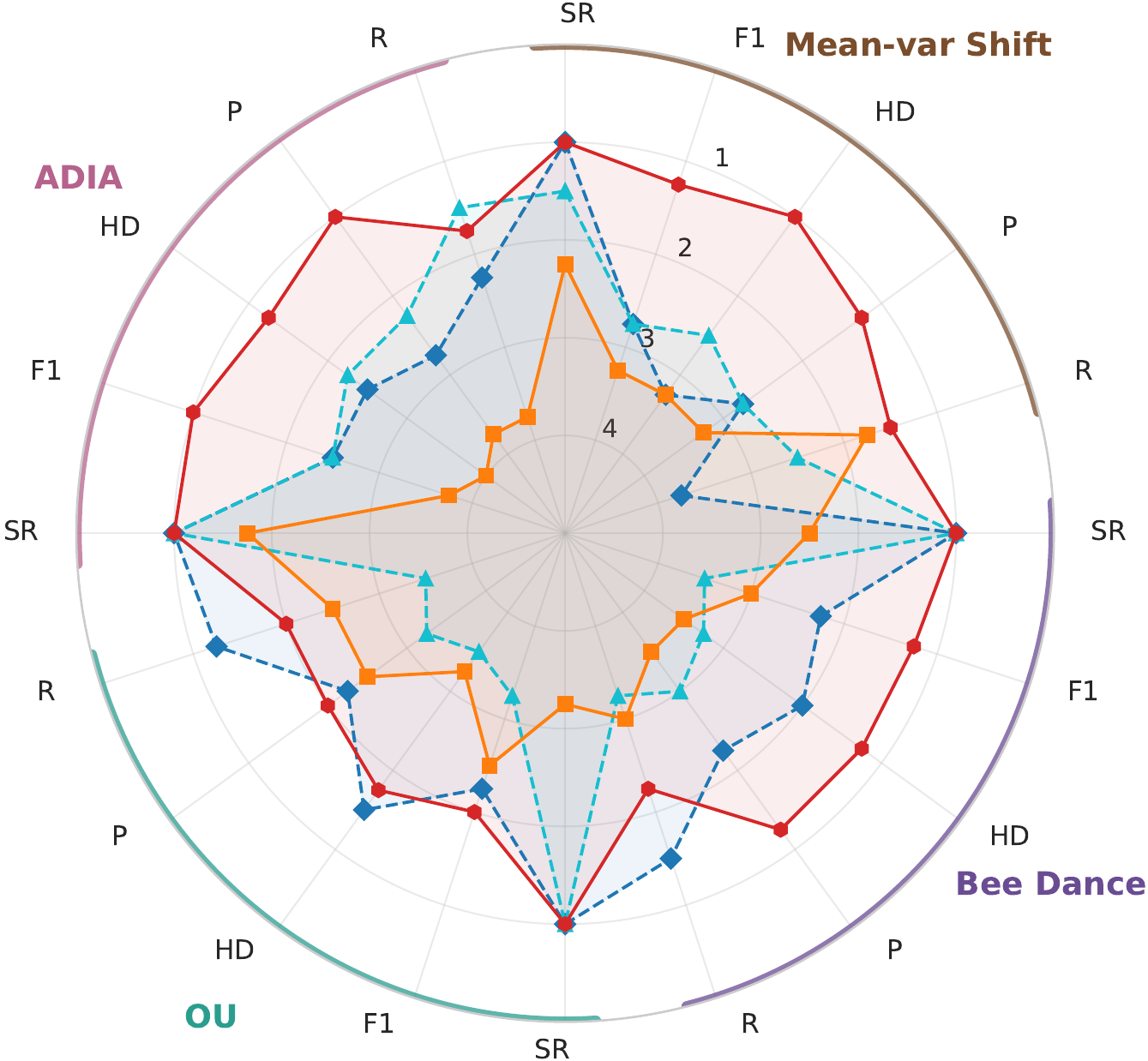}
  \caption{Average agent ranks across four LLM backbones for each dataset–metric pair. Lower average ranks, plotted farther outward, indicate better performance.}
  \label{fig:radar}
\end{figure}

Table~\ref{tab:cpd_all} compares EvoTS-Agent with three representative LLM-based agents across four change-point detection benchmarks using four backbone LLMs. Overall, EvoTS-Agent consistently achieves the strongest or highly competitive performance across datasets while maintaining a 100\% execution success rate for all backbone models. Figure~\ref{fig:radar} further summarizes the results by reporting the average rank of each agent across all backbone LLMs for every dataset–metric pair. Across all four benchmarks, EvoTS-Agent consistently achieves the best or near-best overall ranks, demonstrating strong performance across multiple evaluation metrics rather than on F1 alone. 

For the OU-based financial benchmark, EvoTS-Agent obtains the best GPT-4o performance and remains competitive across other backbone LLMs. Notably, ResearchAgent occasionally achieves high F1 under Sonnet models but suffers from severe execution failures, with success rates dropping to only 33.3\% for Sonnet-4.6 and Sonnet-5. In contrast, EvoTS-Agent consistently completes every run successfully, demonstrating that using the model bank and explicit search strategy substantially improves reliability.

On the synthetic Mean-Variance Shift dataset, EvoTS-Agent achieves the highest F1 score for GPT-4o, GPT-5.4, and Sonnet-5, while remaining competitive for Sonnet-4.6. The improvements are accompanied by substantially lower Hausdorff distances and consistently higher precision, indicating that the proposed evolutionary search improves not only detection accuracy but also boundary localization. The performance gains are particularly pronounced for GPT-5.4, where the F1 score increases from 0.568 (TS-Agent) and 0.562 (DS-Agent) to 0.833.

\begin{figure*}[h!]
    \centering
    \includegraphics[width=\linewidth]{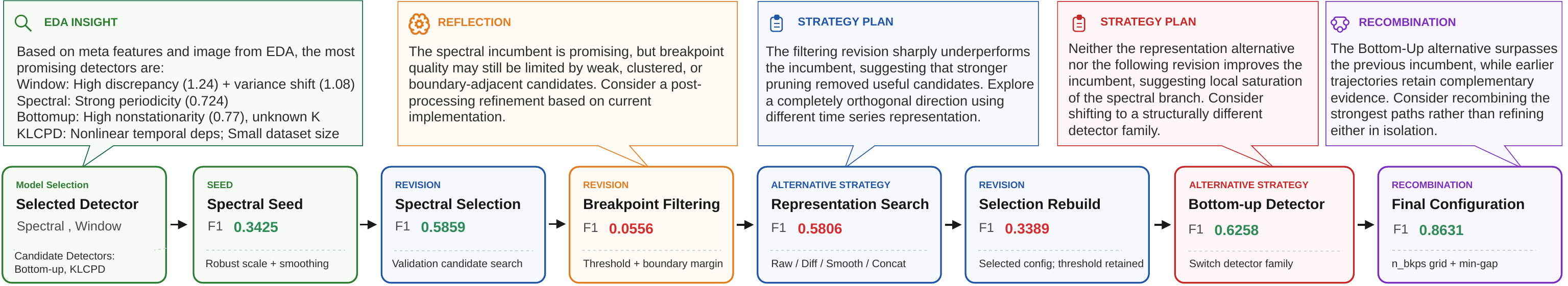}
    \caption{A case study of our EvoTS-Agent on the Bee Dance dataset. The F1 scores shown are evaluated on the validation set.}
    \label{fig:case-study}
\end{figure*}

The largest improvements occur on the ADIA benchmark. EvoTS-Agent achieves the highest F1 score across all four backbone LLMs while substantially reducing the Hausdorff distance. For example, under GPT-4o, the Hausdorff distance decreases from 212.0 (TS-Agent) and 130.6 (DS-Agent) to 14.8, while the F1 score improves from 0.235 and 0.461 to 0.500. These results indicate that EvoTS-Agent effectively adapts to challenging multi-break scenarios involving successive volatility regime changes, where the search strategy benefits from exploring alternative detection models rather than repeatedly refining a single solution.

On the real-world Bee Dance benchmark, the performance differences among the agents are more modest than on the synthetic datasets. TS-Agent achieves the higher F1 score and the lower Hausdorff distance under GPT-4o, while EvoTS-Agent achieves the better F1 score and Hausdorff under GPT-5.4, Sonnet-4.6, and Sonnet-5. These results suggest that the proposed trajectory evolution strategy generalizes well across different backbone LLMs, although its advantage is less pronounced on this challenging real-world dataset.

Although EvoTS-Agent does not attain the highest mean F1 under every backbone LLM, it achieves the best observed individual-run F1 on each benchmark. While Table~\ref{tab:cpd_all} reports the mean and standard deviation over three runs, EvoTS-Agent reaches peak F1 scores of 0.877 on Mean-Variance Shift, 0.682 on Bee Dance, 0.794 on OU, and 1.000 on ADIA. This suggests that the proposed trajectory evolution strategy is more capable of discovering highly effective detection pipelines than competing agentic workflows.

\subsection{Comparison with Individual Detectors}
\input{tables/each_model_performance}

Table~\ref{tab:direct} reports the performance of individual detectors from our model bank using manually specified configurations on the Bee-Dance dataset. The results demonstrate that different detectors exhibit distinct strengths and weaknesses, reflecting the fundamentally different assumptions they make about regime stability. For example, KL-CPD achieves the highest F1 score (0.570), followed closely by Bottom-Up (0.565) and PELT (0.539), while several other detectors perform substantially worse. These observations indicate that no single detector is universally optimal, and that the effectiveness of a method depends on the characteristics of the underlying structural changes. The substantial performance variation further highlights the importance of selecting an appropriate detector and optimizing it for the target dataset.

Starting from the same model bank, EvoTS-Agent with GPT-5.4 improves the F1 score to 0.635. Unlike selecting a fixed detector a priori, EvoTS-Agent adaptively chooses candidate models guided by curated EDA and iteratively refines the detection pipeline through validation-guided evolution, including modifications to data representation, detector configuration, hyperparameters, and post-processing. The improvement therefore arises not simply from having access to multiple detection algorithms, but from the agent's ability to identify and evolve the detector configuration that best matches the statistical characteristics of the target dataset.


\subsection{Ablation Studies}
\input{tables/ablation_study}

Table~\ref{tab:ablation_beedance_gpt54} evaluates the contributions of Alternative Strategy and Recombination on the Bee-Dance dataset using GPT-5.4. The complete EvoTS-Agent achieves the highest F1 score of 0.635, compared with 0.578 without Alternative Strategy and 0.575 without Recombination. Removing Alternative Strategy reduces both precision and recall, indicating that repeatedly refining the incumbent can restrict the search to a suboptimal modeling direction. Removing Recombination causes a particularly clear reduction in recall, from 0.671 to 0.602, suggesting that the final synthesis step helps consolidate complementary evidence discovered by different trajectories. Although the variant without Recombination obtains a  lower Hausdorff distance, its lower F1 and recall indicate weaker overall change-point coverage. Taken together, these results show that the two operators both improve the precision–recall balance and that combining them produces the strongest overall detection performance.

\subsection{Case Study}

Figure~\ref{fig:case-study} illustrates a representative evolutionary trajectory produced by EvoTS-Agent on the Bee-Dance dataset. Starting from an EDA-selected spectral detector, the Revision operator substantially improves the validation F1 score through iterative refinement. Once further revisions no longer yield meaningful gains, EvoTS-Agent detects stagnation and activates Alternative Strategy, finally switching to the Bottom-Up detector family. This transition produces another significant improvement, demonstrating the benefit of exploring detectors with different assumptions about change-point structure rather than repeatedly refining a single approach. In the final iteration, Recombination integrates complementary improvements accumulated from the two best-performing trajectories to produce the best validation F1 score of 0.8631. Overall, this example illustrates how EvoTS-Agent balances exploitation and exploration throughout the evolutionary process.


%% file: tables/agent_comparison.tex
\begin{table*}[h]
\centering
\caption{Time series change point detection performance on four benchmark datasets. Each metric is averaged over three runs. The best result for each LLM (per column) across different agents is bolded.}
\label{tab:cpd_all}
\resizebox{\textwidth}{!}{%
\begin{tabular}{cccccccccccccccccccccc}
\toprule
\multirow{2}[3]{*}{\textbf{Dataset}} & \multirow{2}[3]{*}{\textbf{Model}} & \multicolumn{4}{c}{\textbf{F1} $\uparrow$} & \multicolumn{4}{c}{\textbf{Hausdorff Distance} $\downarrow$} & \multicolumn{4}{c}{\textbf{Precision} $\uparrow$} & \multicolumn{4}{c}{\textbf{Recall} $\uparrow$} & \multicolumn{4}{c}{\textbf{Success Rate} (\%) $\uparrow$} \\
\cmidrule(lr){3-6}\cmidrule(lr){7-10}\cmidrule(lr){11-14}\cmidrule(lr){15-18}\cmidrule(lr){19-22}
& & GPT-4o & GPT-5.4 & Sonnet-4.6 & Sonnet-5 & GPT-4o & GPT-5.4 & Sonnet-4.6 & Sonnet-5 & GPT-4o & GPT-5.4 & Sonnet-4.6 & Sonnet-5 & GPT-4o & GPT-5.4 & Sonnet-4.6 & Sonnet-5 & GPT-4o & GPT-5.4 & Sonnet-4.6 & Sonnet-5 \\
\midrule
\multirow{4}{*}{OU-based Dataset}
& \textsf{TS-Agent} &
$0.728^{\pm 0.02}$ & $\mathbf{0.696}^{\pm 0.04}$ &
$0.633^{\pm 0.08}$ & $0.680^{\pm 0.05}$ &
$\mathbf{47.67}^{\pm 0.58}$ & $\mathbf{48.94}^{\pm 1.99}$ &
$53.89^{\pm 3.52}$ & $\mathbf{46.44}^{\pm 1.54}$ &
$0.708^{\pm 0.01}$ & $\mathbf{0.666}^{\pm 0.05}$ &
$0.570^{\pm 0.12}$ & $0.676^{\pm 0.03}$ &
$\mathbf{0.889}^{\pm 0.00}$ & $\mathbf{0.889}^{\pm 0.00}$ &
$\mathbf{0.889}^{\pm 0.00}$ & $0.815^{\pm 0.13}$ &
\textbf{100} & \textbf{100} & \textbf{100} & \textbf{100} \\
& \textsf{DS-Agent} &
$0.631^{\pm 0.28}$ & $0.671^{\pm 0.11}$ &
$0.502^{\pm 0.12}$ & $0.583^{\pm 0.09}$ &
$57.83^{\pm 17.03}$ & $64.67^{\pm 20.30}$ &
$58.00^{\pm 12.60}$ & $64.28^{\pm 8.77}$ &
$0.620^{\pm 0.32}$ & $0.658^{\pm 0.14}$ &
$0.431^{\pm 0.11}$ & $0.671^{\pm 0.01}$ &
$0.741^{\pm 0.26}$ & $0.815^{\pm 0.08}$ &
$0.833^{\pm 0.00}$ & $0.648^{\pm 0.20}$ &
\textbf{100} & \textbf{100} & \textbf{100} & \textbf{100} \\
& \textsf{ResearchAgent} &
$0.473^{\pm 0.31}$ & N/A &
$\mathbf{0.794}^{\pm 0.00}$ &
$\mathbf{0.794}^{\pm 0.00}$ &
$80.22^{\pm 24.81}$ & N/A &
$48.17^{\pm 0.00}$ & $47.83^{\pm 0.00}$ &
$0.484^{\pm 0.33}$ & N/A &
$\mathbf{0.806}^{\pm 0.00}$ & $\mathbf{0.806}^{\pm 0.00}$ &
$0.574^{\pm 0.37}$ & N/A &
$\mathbf{0.889}^{\pm 0.00}$ & $\mathbf{0.889}^{\pm 0.00}$ &
\textbf{100} & 0 & 33.3 & 33.3 \\
& \textsf{EvoTS-Agent} &
$\mathbf{0.767}^{\pm 0.05}$ & $0.626^{\pm 0.18}$ &
$0.671^{\pm 0.11}$ & $0.717^{\pm 0.13}$ &
$47.89^{\pm 0.48}$ & $57.56^{\pm 13.88}$ &
$\mathbf{47.78}^{\pm 4.10}$ & $47.33^{\pm 1.17}$ &
$\mathbf{0.769}^{\pm 0.06}$ & $0.552^{\pm 0.24}$ &
$0.625^{\pm 0.16}$ & $0.731^{\pm 0.13}$ &
$\mathbf{0.889}^{\pm 0.00}$ & $\mathbf{0.889}^{\pm 0.06}$ &
$0.852^{\pm 0.03}$ & $0.796^{\pm 0.16}$ &
\textbf{100} & \textbf{100} & \textbf{100} & \textbf{100} \\
\midrule
\multirow{4}{*}{Mean-variance Shift}
& \textsf{TS-Agent} &
$0.355^{\pm 0.18}$ &
$0.568^{\pm 0.13}$ &
$0.608^{\pm 0.04}$ &
$0.789^{\pm 0.13}$ &
$128.11^{\pm 5.91}$ &
$79.11^{\pm 10.78}$ &
$74.22^{\pm 9.33}$ &
$33.83^{\pm 30.31}$ &
$0.537^{\pm 0.38}$ &
$0.634^{\pm 0.13}$ &
$0.620^{\pm 0.08}$ &
$0.833^{\pm 0.14}$ &
$0.315^{\pm 0.07}$ &
$0.556^{\pm 0.10}$ &
$0.611^{\pm 0.00}$ &
$0.759^{\pm 0.13}$ &
\textbf{100} & \textbf{100} & \textbf{100} & \textbf{100} \\
& \textsf{DS-Agent} &
$0.731^{\pm 0.14}$ &
$0.562^{\pm 0.22}$ &
$0.748^{\pm 0.06}$ &
$0.686^{\pm 0.06}$ &
$41.06^{\pm 37.05}$ &
$58.89^{\pm 41.17}$ &
$54.28^{\pm 19.89}$ &
$41.42^{\pm 27.22}$ &
$0.796^{\pm 0.11}$ &
$0.568^{\pm 0.20}$ &
$0.806^{\pm 0.10}$ &
$0.685^{\pm 0.01}$ &
$0.704^{\pm 0.13}$ &
$0.630^{\pm 0.26}$ &
$0.722^{\pm 0.06}$ &
$0.736^{\pm 0.14}$ &
\textbf{100} & \textbf{100} & \textbf{100} & 66.7 \\
& \textsf{ResearchAgent} &
$0.181^{\pm 0.08}$ &
$0.282^{\pm 0.00}$ &
$\mathbf{0.794}^{\pm 0.13}$ &
$0.598^{\pm 0.31}$ &
$82.94^{\pm 31.27}$ &
$90.83^{\pm 0.00}$ &
$37.61^{\pm 36.85}$ &
$59.67^{\pm 40.07}$ &
$0.118^{\pm 0.06}$ &
$0.183^{\pm 0.00}$ &
$\mathbf{0.861}^{\pm 0.10}$ &
$0.577^{\pm 0.48}$ &
$0.704^{\pm 0.42}$ &
$0.611^{\pm 0.00}$ &
$\mathbf{0.759}^{\pm 0.13}$ &
$\mathbf{0.889}^{\pm 0.16}$ &
\textbf{100} & 33.3 & \textbf{100} & 66.7 \\
& \textsf{EvoTS-Agent} &
$\mathbf{0.770}^{\pm 0.12}$ &
$\mathbf{0.833}^{\pm 0.06}$ &
$0.759^{\pm 0.09}$ &
$\mathbf{0.856}^{\pm 0.02}$ &
$\mathbf{38.89}^{\pm 25.93}$ &
$\mathbf{17.11}^{\pm 1.35}$ &
$\mathbf{32.22}^{\pm 13.31}$ &
$\mathbf{18.17}^{\pm 3.18}$ &
$\mathbf{0.815}^{\pm 0.14}$ &
$\mathbf{0.889}^{\pm 0.05}$ &
$0.843^{\pm 0.07}$ &
$\mathbf{0.917}^{\pm 0.00}$ &
$\mathbf{0.741}^{\pm 0.12}$ &
$\mathbf{0.796}^{\pm 0.06}$ &
$0.722^{\pm 0.10}$ &
$0.815^{\pm 0.03}$ &
\textbf{100} & \textbf{100} & \textbf{100} & \textbf{100} \\
\midrule
\multirow{4}{*}{ADIA Dataset}
& \textsf{TS-Agent} &
$0.235^{\pm 0.09}$ &
$0.398^{\pm 0.03}$ &
$0.417^{\pm 0.00}$ &
$0.370^{\pm 0.16}$ &
$212.00^{\pm 1.74}$ & $36.78^{\pm 3.66}$ &
$34.67^{\pm 0.00}$ & $83.33^{\pm 101.95}$ &
$0.199^{\pm 0.12}$ & $0.361^{\pm 0.05}$ &
$0.389^{\pm 0.00}$ & $0.343^{\pm 0.18}$ &
$0.333^{\pm 0.00}$ & $\mathbf{0.500}^{\pm 0.00}$ &
$0.500^{\pm 0.00}$ & $0.444^{\pm 0.10}$ &
\textbf{100} & \textbf{100} & \textbf{100} & \textbf{100} \\
& \textsf{DS-Agent} &
$0.461^{\pm 0.49}$ &
$0.389^{\pm 0.19}$ &
$\mathbf{0.667}^{\pm 0.29}$ &
$0.153^{\pm 0.01}$ &
$130.56^{\pm 104.41}$ & $101.67^{\pm 150.11}$ &
$\mathbf{10.00}^{\pm 0.00}$ & $152.22^{\pm 26.17}$ &
$0.409^{\pm 0.52}$ & $0.389^{\pm 0.19}$ &
$\mathbf{0.667}^{\pm 0.29}$ & $0.087^{\pm 0.01}$ &
$\mathbf{0.889}^{\pm 0.19}$ & $0.389^{\pm 0.19}$ &
$\mathbf{0.667}^{\pm 0.29}$ & $\mathbf{0.722}^{\pm 0.25}$ &
\textbf{100} & \textbf{100} & \textbf{100} & \textbf{100} \\
& \textsf{ResearchAgent} &
$0.074^{\pm 0.08}$ & N/A &
$0.309^{\pm 0.20}$ &
$0.278^{\pm 0.19}$ &
$238.39^{\pm 139.96}$ & N/A &
$68.89^{\pm 93.92}$ &
$188.56^{\pm 149.73}$ &
$0.044^{\pm 0.05}$ & N/A &
$0.295^{\pm 0.23}$ &
$0.278^{\pm 0.19}$ &
$0.222^{\pm 0.25}$ & N/A &
$0.500^{\pm 0.17}$ &
$0.278^{\pm 0.19}$ &
\textbf{100} & 0 & \textbf{100} & \textbf{100} \\
& \textsf{EvoTS-Agent} &
$\mathbf{0.500}^{\pm 0.00}$ &
$\mathbf{0.444}^{\pm 0.10}$ &
$\mathbf{0.667}^{\pm 0.29}$ &
$\mathbf{0.444}^{\pm 0.10}$ &
$\mathbf{14.83}^{\pm 0.00}$ & $\mathbf{16.17}^{\pm 2.31}$ &
$12.94^{\pm 3.42}$ & $\mathbf{14.72}^{\pm 4.17}$ &
$\mathbf{0.500}^{\pm 0.00}$ & $\mathbf{0.444}^{\pm 0.10}$ &
$\mathbf{0.667}^{\pm 0.29}$ & $\mathbf{0.444}^{\pm 0.10}$ &
$0.500^{\pm 0.00}$ & $0.444^{\pm 0.10}$ &
$\mathbf{0.667}^{\pm 0.29}$ & $0.444^{\pm 0.10}$ &
\textbf{100} & \textbf{100} & \textbf{100} & \textbf{100} \\
\midrule
\multirow{4}{*}{Bee Dance}
& \textsf{TS-Agent} &
$\mathbf{0.561}^{\pm 0.10}$ &
$0.576^{\pm 0.13}$ &
$0.546^{\pm 0.10}$ &
$0.520^{\pm 0.05}$ &
$\mathbf{26.83}^{\pm 7.25}$ &
$40.06^{\pm 11.72}$ &
$33.72^{\pm 3.91}$ &
$24.44^{\pm 2.08}$ &
$0.498^{\pm 0.11}$ &
$0.544^{\pm 0.15}$ &
$\mathbf{0.564}^{\pm 0.10}$ &
$0.404^{\pm 0.07}$ &
$\mathbf{0.699}^{\pm 0.20}$ &
$\mathbf{0.685}^{\pm 0.29}$ &
$0.556^{\pm 0.10}$ &
$\mathbf{0.792}^{\pm 0.15}$ &
\textbf{100} & \textbf{100} & \textbf{100} & \textbf{100} \\
& \textsf{DS-Agent} &
$0.127^{\pm 0.10}$ &
$0.383^{\pm 0.18}$ &
$0.271^{\pm 0.13}$ &
$0.522^{\pm 0.06}$ &
$73.78^{\pm 70.88}$ &
$51.61^{\pm 45.23}$ &
$72.94^{\pm 27.41}$ &
$24.11^{\pm 1.99}$ &
$0.116^{\pm 0.07}$ &
$0.346^{\pm 0.03}$ &
$0.454^{\pm 0.11}$ &
$0.442^{\pm 0.02}$ &
$0.519^{\pm 0.42}$ &
$0.681^{\pm 0.48}$ &
$0.208^{\pm 0.13}$ &
$0.681^{\pm 0.19}$ &
\textbf{100} & \textbf{100} & \textbf{100} & \textbf{100} \\
& \textsf{ResearchAgent} &
$0.225^{\pm 0.00}$ &
N/A &
$0.369^{\pm 0.29}$ &
$0.526^{\pm 0.11}$ &
$52.17^{\pm 0.00}$ &
N/A &
$56.11^{\pm 55.92}$ &
$24.67^{\pm 1.50}$ &
$0.306^{\pm 0.00}$ &
N/A &
$0.300^{\pm 0.20}$ &
$0.427^{\pm 0.15}$ &
$0.181^{\pm 0.00}$ &
N/A &
$0.560^{\pm 0.47}$ &
$0.778^{\pm 0.09}$ &
33.3 & 0 & \textbf{100} & \textbf{100} \\
& \textsf{EvoTS-Agent} &
$0.496^{\pm 0.06}$ &
$\mathbf{0.635}^{\pm 0.04}$ &
$\mathbf{0.576}^{\pm 0.05}$ &
$\mathbf{0.543}^{\pm 0.02}$ &
$29.67^{\pm 6.29}$ &
$\mathbf{37.67}^{\pm 11.05}$ &
$\mathbf{31.06}^{\pm 5.10}$ &
$\mathbf{23.67}^{\pm 2.24}$ &
$\mathbf{0.509}^{\pm 0.08}$ &
$\mathbf{0.660}^{\pm 0.16}$ &
$0.560^{\pm 0.15}$ &
$\mathbf{0.461}^{\pm 0.06}$ &
$0.532^{\pm 0.14}$ &
$0.671^{\pm 0.13}$ &
$\mathbf{0.736}^{\pm 0.18}$ &
$0.713^{\pm 0.17}$ &
\textbf{100} & \textbf{100} & \textbf{100} & \textbf{100} \\
\bottomrule
\end{tabular}%
}
\end{table*}

%% file: tables/each_model_performance.tex
\begin{table}[t]
  \centering
  \caption{Direct results of models in Model Bank on the Bee Dance dataset.}
  \label{tab:direct}

  \resizebox{\columnwidth}{!}{%
    \begin{tabular}{lcccc}
      \toprule
      \textbf{Model}
      & \textbf{F1} $\uparrow$
      & \textbf{Hausdorff Distance} $\downarrow$
      & \textbf{Precision} $\uparrow$
      & \textbf{Recall} $\uparrow$ \\
      \midrule

      \textsf{PELT}
      & $0.539 $
      & 26.50
      & $0.393 $
      & $0.875 $ \\

      \textsf{BottomUp}
      & $0.565 $
      & $26.67 $
      & $0.408 $
      & $0.931 $ \\

      \textsf{Window}
      & $0.523 $
      & $36.50 $
      & $0.519 $
      & $0.750 $ \\

      \textsf{ChangeForest-RF}
      & $0.275 $
      & $31.28 $
      & $0.160 $
      & \textbf{1.000} \\

      \textsf{ChangeForest-KNN}
      & $0.310 $
      & $28.89 $
      & $0.187 $
      & $0.944 $ \\

      \textsf{BOCD}
      & $0.501 $
      & \textbf{24.33}
      & $0.351 $
      & $0.903 $ \\

      \textsf{KLCPD}
      & \textbf{0.570} 
      & $43.11 $
      & $0.694 $
      & $0.532 $ \\

      \textsf{Spectral}
      & $0.483 $
      & $64.00 $
      & \textbf{0.917}
      & $0.333 $ \\

      \bottomrule
    \end{tabular}%
  }

\end{table}

%% file: tables/ablation_study.tex







\begin{table}[t]
\centering
\caption{Ablation study on the Bee-Dance dataset using GPT-5.4.
Results are reported as mean $\pm$ standard deviation over three runs.
The best result in each column is bolded.}
\label{tab:ablation_beedance_gpt54}

\resizebox{\columnwidth}{!}{%
\begin{tabular}{lcccc}
\toprule
\textbf{Model}
& \textbf{F1} $\uparrow$
& \textbf{Hausdorff Distance} $\downarrow$
& \textbf{Precision} $\uparrow$
& \textbf{Recall} $\uparrow$ \\
\midrule

\textsf{EvoTS-Agent}
& $\mathbf{0.635}^{\pm 0.04}$
& $37.67^{\pm 11.05}$
& $\mathbf{0.660}^{\pm 0.16}$
& $\mathbf{0.671}^{\pm 0.13}$ \\

\textsf{EvoTS-Agent w/o Alternative}
& $0.578^{\pm 0.06}$
& $29.67^{\pm 4.77}$
& $0.587^{\pm 0.12}$
& $0.639^{\pm 0.18}$ \\

\textsf{EvoTS-Agent w/o Recombination}
& $0.575^{\pm 0.12}$
& $\mathbf{28.61}^{\pm 3.08}$
& $0.616^{\pm 0.10}$
& $0.602^{\pm 0.13}$ \\

\bottomrule
\end{tabular}%
}

\end{table}

%% file: sec/6_conclusion.tex
\section{Conclusion \& Limitation}

We presented EvoTS-Agent, a self-evolving LLM agent for financial time-series change-point detection. Unlike conventional LLM agents that repeatedly refine a single solution or follow a fixed workflow, EvoTS-Agent performs validation-guided trajectory evolution by maintaining executable experiment trajectories and adapting its search strategy according to empirical feedback. Through curated exploratory data analysis, a diverse change-point detection model bank, and validation-guided trajectory evolution, the agent automatically selects, refines, and synthesizes detection pipelines. Experiments on four benchmark datasets demonstrate that EvoTS-Agent consistently achieves strong change-point detection performance across multiple backbone LLMs while maintaining a 100\% execution success rate. Further analysis shows that the gain arises not only from access to diverse detection algorithms, but also from the ability to adaptively select and evolve the most appropriate detection pipeline for the statistical characteristics of each dataset. 

One limitation of EvoTS-Agent is its dependence on labeled validation data. While the underlying change-point detectors are unsupervised, the agent requires validation annotations to determine which experimental trajectories should be retained during evolution. In practical deployments, obtaining such labels may require expert annotation or historical ground-truth events. Developing reliable label-efficient or label-free optimization strategies is therefore an important avenue for future research.

%% file: sec/7_acknowledge.tex
\begin{acks}
LJ, JL and LS acknowledge the support of the UKRI Prosperity Partnership Scheme (FAIR) under the EPSRC Grant EP/V056883/1 and The Alan Turing Institute.
XX, YB, YA, and AT are supported by the National Research Foundation, Singapore, under its CyberSG R\&D Programme (CRPO Award No: CRPO-GC5-NUS-006), the Ministry of Education, Singapore, under its MOE AcRF TIER 1 Grant (T1 251RES2517), and the National University of Singapore, School of Computing Seed Fund. 
HN is supported in part by the EPSRC Program Grant [Grant No. UKRI1010] entitled ``High order mathematical and computational infrastructure for streamed data that enhance contemporary generative and large language models'' and by the SURE-AI Centre grant 357482, Research Council of Norway.

\end{acks}